\documentclass[letterpaper, 10 pt, conference]{ieeeconf}  % Comment this line out if you need a4paper

\IEEEoverridecommandlockouts                              % This command is only needed if 
\usepackage{times}
\usepackage{epsfig}
\usepackage{graphicx}
\usepackage{amsmath}
\usepackage{amssymb}
\usepackage{booktabs,subcaption,amsfonts,dcolumn}
\usepackage{listings}
\usepackage{color}
\usepackage[dvipsnames]{xcolor}
\usepackage{tablefootnote}

\definecolor{codegreen}{rgb}{0,0.6,0}
\definecolor{codegray}{rgb}{0.5,0.5,0.5}
\definecolor{codepurple}{rgb}{0.58,0,0.82}
\definecolor{backcolour}{rgb}{0.98,0.98,0.98}
\usepackage{multirow}
\usepackage{graphicx}
\usepackage{hyperref}
\usepackage{multirow}
\usepackage{tabularx}
\usepackage{booktabs,subcaption,amsfonts,dcolumn}
\usepackage{listings}
\usepackage{color}
\definecolor{backcolour}{rgb}{0.95, 0.95, 0.96}
\usepackage[dvipsnames]{xcolor}
\usepackage{tablefootnote}
\newcommand{\pva}{PVTransformer}

\usepackage{graphics} % for pdf, bitmapped graphics files
\usepackage{epsfig} % for postscript graphics files
\usepackage{mathptmx} % assumes new font selection scheme installed
\usepackage{amsmath} % assumes amsmath package installed
\usepackage{amssymb}  % assumes amsmath package installed

\title{\pva{}: Point-to-Voxel Transformer for Scalable 3D Object Detection}

\author{Zhaoqi Leng$^{1}$, Pei Sun$^{1}$, Tong He$^{1}$, Dragomir Anguelov$^{1}$, and Mingxing Tan$^{1}$
\thanks{$^{1}$Waymo Research}%
}

\begin{document}

\maketitle

%%%%%%%%%%%%%%%%%%%%%%%%%%%%%%%%%%%%%%%%%%%%%%%%%%%%%%%%%%%%%%%%%%%%%%%%%%%%%%%%

\begin{abstract}
3D object detectors for point clouds often rely on a pooling-based PointNet~\cite{qi2017pointnet} to encode sparse points into grid-like voxels or pillars. In this paper, we identify that the common PointNet design introduces an information bottleneck that limits 3D object detection accuracy and scalability. To address this limitation, we propose \pva{}: a transformer-based point-to-voxel architecture for 3D detection. Our key idea is to replace the PointNet pooling operation with an attention module, leading to a better point-to-voxel aggregation function. Our design respects the permutation invariance of sparse 3D points while being more expressive than the pooling-based PointNet. Experimental results show our \pva{} achieves much better performance compared to the latest 3D object detectors. On the widely used Waymo Open Dataset, our \pva{} achieves state-of-the-art 76.5 mAPH L2, outperforming the prior art of SWFormer~\cite{sun2022swformer} by +1.7 mAPH L2.
\end{abstract}
\section{Introduction}
3D object detection for autonomous driving in urban environments requires processing a large amount of sparse and unordered points scattered across an open three-dimensional space. To manage the irregular distribution of points, existing methods aggregate points into two or three-dimensional voxel representation \cite{zhou2018voxelnet}, utilizing a PointNet-type feature encoder \cite{qi2017pointnet} to aggregate point features to voxels, which is then followed by a backbone and detection heads. However, the existing point architecture is often overlooked and limited by its minimalist design, i.e., a few fully connected layers followed by a max pooling layer. The key to PointNet-type modules, as highlighted in the original paper \cite{qi2017pointnet}, lies in the max pooling layer, which extracts information from unordered points and serves as an aggregation function. 
Despite the utilization of numerous fully connected layers for feature extraction, the features of all points within a voxel are combined through a simple pooling layer. 
\begin{figure}
    \centering
    \includegraphics[width=0.8\linewidth]{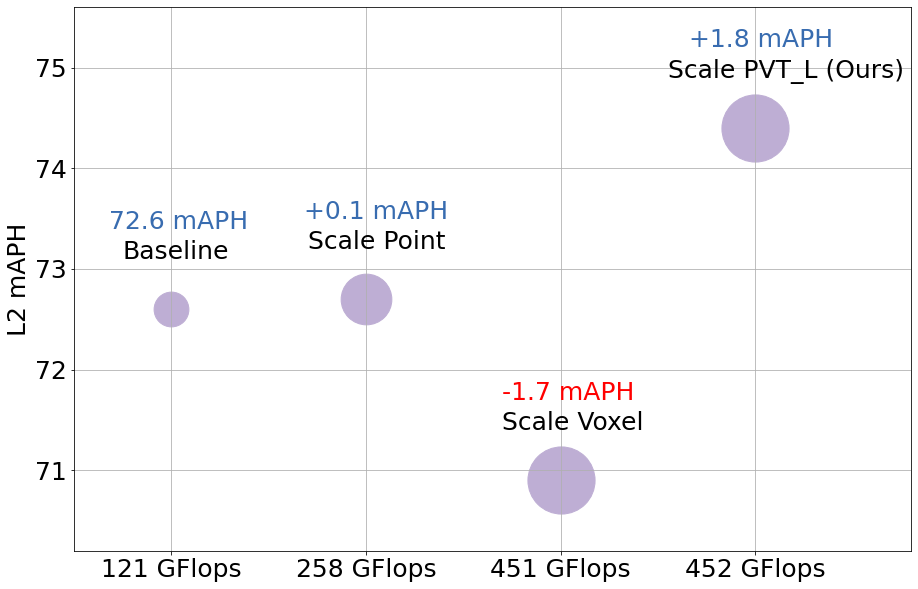}
    \caption{\textbf{\pva{} (PVT) as a scalable architecture.} \pva{} addresses the pooling bottleneck in previous voxel-based 3D detectors and demonstrates better scalability compared to scaling PointNet (Scale Point) and voxel architectures (Scale Voxel). The size of each point represents the model Flops. More details are shown in \autoref{fig:point_scaling}, \ref{fig:backbone_scaling}.
    }
    \vspace{-4mm}
    \label{fig:teaser}
\end{figure}
We observe that the vanilla pooling operation in 3D object detection introduces an information bottleneck, hindering the performance of modern 3D object detectors. Unlike the standard 2D max-pooling in image recognition, which operates on a limited set of pixels, the point-voxel pooling layer in 3D detectors must aggregate many unordered points. For instance, in the Waymo Open Dataset \cite{sun2020scalability}, it is common to find over 100 points in a single $0.32\,$m $\times 0.32\,$m voxel, pooled into a single voxel feature vector. This results in a significant loss of information from the point features after the pooling layer. 

To address the limitations of the pooling-based PointNet architecture, we introduce \textbf{\pva{}}, a new attention-point-voxel architecture for 3D object detection based on the Transformer \cite{vaswani2017attention}. The aim of \pva{} is to mitigate the information bottleneck introduced by the pooling operation in modern 3D object detectors, by end-to-end learning a point-to-voxel encoding function via an attention module. In \pva{}, each point in a voxel is considered a token, and a single query vector is used to query all the point tokens, thereby aggregating and encoding all point features within a voxel into a single voxel feature vector. The attention-based aggregation module in \pva{} serves as a set operator, maintaining permutation invariance, but is more expressive than max pooling. Notably, unlike other Transformer-based point networks such as Point Transformer~\cite{zhao2021point}, which use pooling to aggregate points, \pva{} is designed to \textit{learn} a feature aggregation function without the need for a heuristic pooling.

We evaluate our \pva{} on the Waymo Open Dataset, the largest public 3D point cloud dataset  \cite{sun2020scalability}. Experimental results show that our \pva{} significantly outperforms previous PointNet-based 3D object detectors, by improving point-to-voxel aggregation. Moreover,  \pva{} enables us to scale up the model, achieving a new state-of-the-art 76.1 mAPH L2 and 85.0/84.7 AP L1 for vehicles and pedestrians. Notably, our voxel backbone and loss design are largely based on the prior art of SWFormer~\cite{sun2022swformer}, but our newly proposed point-to-voxel Transformer achieves +1.7 mAHP L2 compared to the baseline SWFormer.

Our contributions can be summarized as:
\begin{itemize}
\item \textbf{New architecture:} Introduction of an attention-based point-voxel architecture, \pva{}, aiming to address the pooling limitation of PointNet.
\item \textbf{Novel scaling studies:} Initiating the exploration of scalability in Transformer-based 3D detector architectures.
\item  \textbf{Extensive studies:} Through extensive architecture search, we show the effectiveness of the proposed \pva{} architecture, which achieves a state-of-the-art 76.5 mAPH L2 on the Waymo Open Dataset. 
\end{itemize}

\section{Related works}
3D object detection has been well explored in previous works, and can be roughly divided into point-based and voxel-based approaches.  Most point-based detectors ~\cite{qi2018frustum,ngiam2019starnet,shi2019pointrcnn}, following seminal works such as PointNet~\cite{qi2017pointnet} and PointNet++ \cite{qi2017pointnet++}, directly operate on the unordered points with fully connected layers to learn per-point features. These models are often not scalable because directly processing large amount of unordered points is expensive \cite{liu2019point}.

On the contrary, voxel-based models, such as VoxelNet~\cite{zhou2018voxelnet} and PointPillars~\cite{lang2019pointpillars}, are generally more scalable by discretizing the sparse point clouds into grids, and thus become increasingly popular in recent research. Voxel-based detectors first convert unordered points into ordered two or three-dimensional voxel grids, by aggregating nearby points based on their coordinates, and then operate on the voxel grid to produce box predictions. The aggregation step typically relies on a simple max pooling layer, which introduces an information bottleneck. Our \pva{} aims to redesign this fundamental pooling-based point aggregation module with a transformer-based learned aggregation function.

 Recent works \cite{Meyer_2019_lasernet,chai2021point, fan2021rangedet,bewley2020range} attempt to directly predict 3D bounding boxes purely based on range images. Meanwhile, multi-view 3D detectors~\cite{zhou2019end, sun2021rsn} leverage multiple views of point clouds, such as points, voxels, and range images. Transformer architecture is also adopted as the backbone \cite{zhao2021point,misra2021end,liu2019point2sequence,yang2019modeling,lee2019set, singlestride21},  as the set operation is naturally suitable for processing sparse LiDAR points or voxels. Unlike those works, \pva{} aims to use the attention module in Transformers to learn a feature aggregation function between the point and voxel representations.

\begin{figure*}[t!]
    \centering
    \vspace{0.2cm}
    \includegraphics[width=0.75\textwidth]{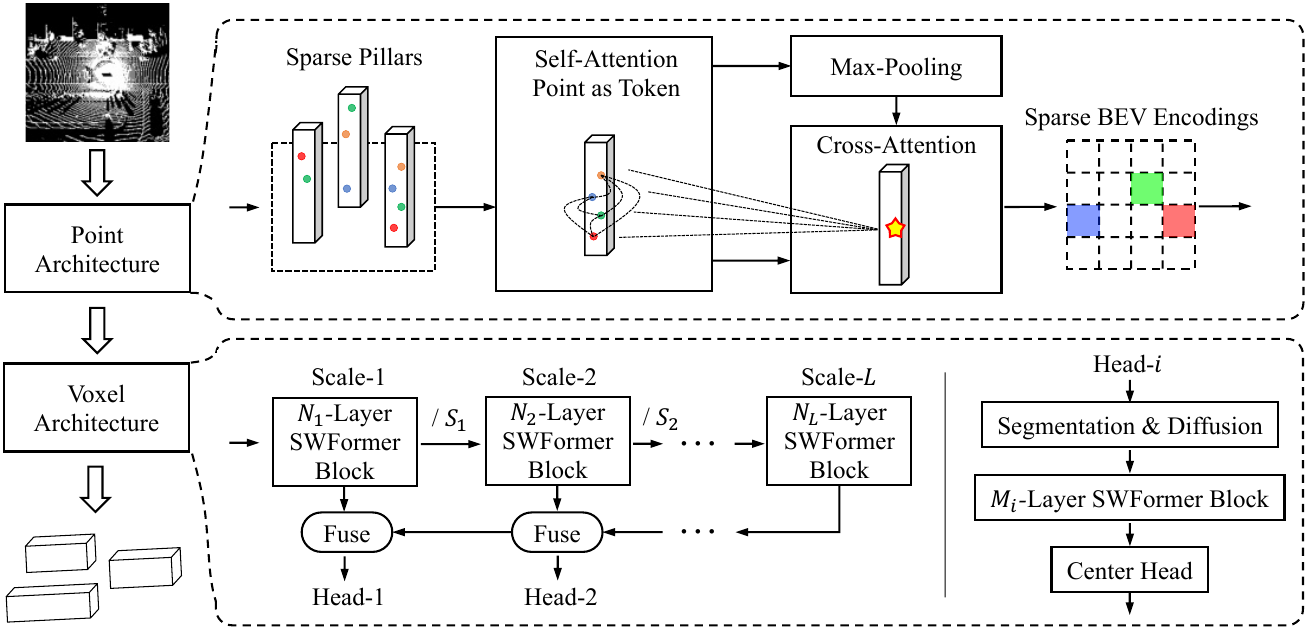}
    \caption{\textbf{Overview of the \pva{} architecture.} The \pva{} architecture contains a point architecture and a voxel architecture. Its novelty lies in the point architecture, substituting PointNet with a novel Transformer design. In the point architecture, points are bucked into pillars, and each is considered as a token. Within a voxel, points undergo a self-attention Transformer followed by a cross-attention Transformer to aggregate point features into voxel features,  details further shown in \autoref{fig:point_arch} (b). The sparse BEV voxel features proceed to the voxel architecture, employing a multi-scale sparse window Transformer (SWFormer Block) \cite{sun2022swformer} for encoding and CenterNet heads for bounding boxes predictions \cite{yin2021centerpoint}.
    }
    \label{fig:arch}
\end{figure*}

\begin{figure}[t!]
    \centering
    \includegraphics[width=0.48\textwidth]{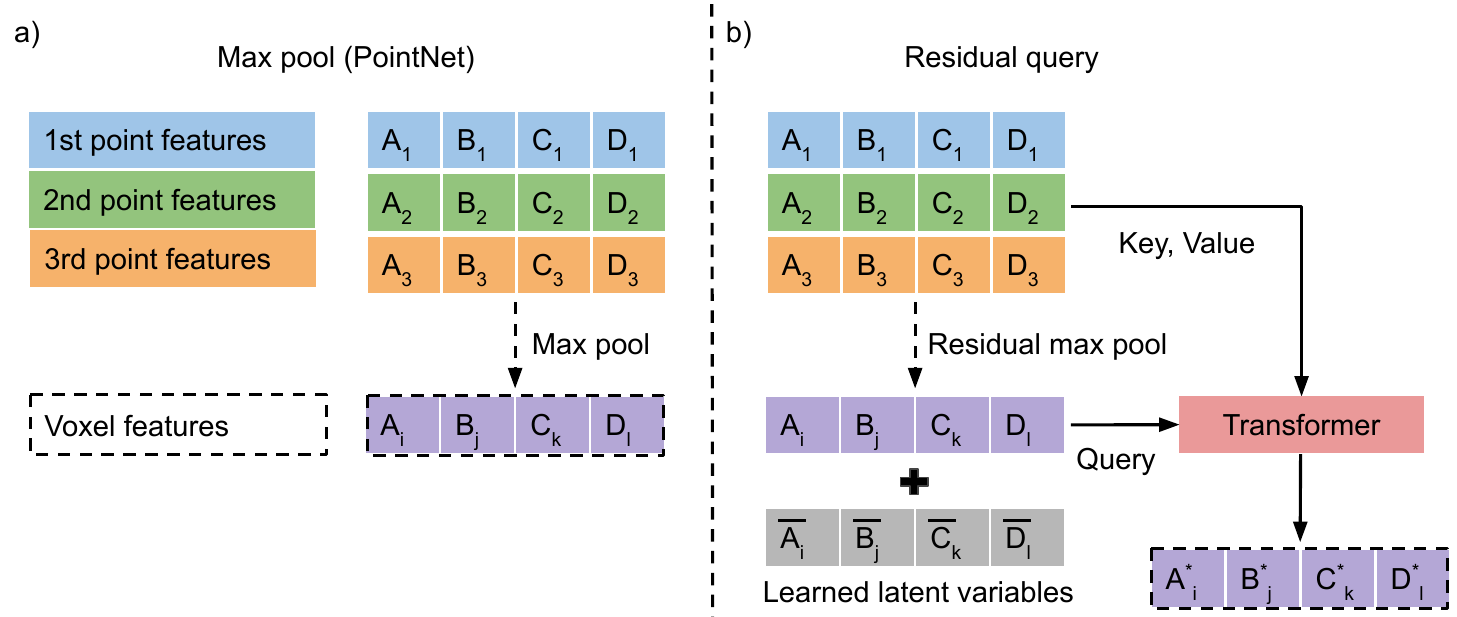}
    \caption{\textbf{Point-to-voxel aggregation in \pva{}.} This module replaces PointNet's max pooling \cite{qi2017pointnet} with a Transformer layer. (a) The vanilla max pooling layer aggregates point features by selecting the element-wise max feature $(A_i, B_j, C_k, D_l)$ from 3 point features to form the voxel features, where $i, j, k, l \in [1, 3]$.  (b) Our proposed residual query uses the sum of max pooled feature and a learnable latent vector $(\overline{A_i}, \overline{B_j}, \overline{C_k}, \overline{D_l})$, i.e. $(A_i + \overline{A_i}, B_i + \overline{B_i}, C_i + \overline{C_i}, D_i + \overline{D_i})$, to query and aggregate point features. 
    }
    \label{fig:point_arch}
    \vspace{-0.5cm}

\end{figure}

\section{\pva{}}
In this section, we provide an overview of our proposed \pva{} architecture for voxel-based 3D detectors. First, we present the conventional network structure of voxel-based 3D detectors. Then, we introduce the details of our \pva{} architecture and its key innovations. Finally, we discuss the scalability of the proposed \pva{}.

\subsection{Voxel-based 3D detectors} \label{sec:pillar3d}
In this work, we use 2D voxel-based (pillar) representation, which is introduced in PointPillars~\cite{lang2019pointpillars}. A typical voxel-based 3D detector starts with raw 3D point clouds and uses a lightweight PointNet to extract per-point features. These point features are then aggregated into individual voxel grids using a max pooling layer, followed by a high-capacity backbone network for final box predictions.

As illustrated in \autoref{fig:point_arch} (a), the pooling layer in PointNet aggregates all the point features, represented as $\mathbb{R}^{n\times d}$ within a single voxel, into a single voxel feature $\mathbb{R}^{1\times d}$. Here, $n$ represents the number of points per voxel and $d$ represents the feature size. On the Waymo Open Dataset, the number of LiDAR points $n$ in a pillar with size $0.32 \,$m $\times 0.32\,$m  commonly exceeds 30, resulting in a $30\times$ compression of information. This aggressive pooling operation causes significant information loss and creates an information bottleneck between point features and voxel features.

To overcome this limitation, we propose to use attention to learn the weighting of point features, instead of the sample max pooling operation. This idea leads to our proposed architecture \pva{}.

\subsection{Point-to-voxel encoding} \label{sec:feature_agg}
A key contribution of our \pva{} architecture is replacing the \textit{non-trainable} pooling operation used in PointNet with a \textit{trainable} Transformer-based aggregation module.  

\textbf{Treating points as tokens}. In \pva{}, each point within a voxel is treated as a token and fed to a Transformer layer. A query vector, learned from all the points within the voxel, is used to encode the point information into a single voxel representation, which we will discuss in detail in the next paragraph. The advantage of this design is that a multi-head attention module can learn a much more expressive point-to-voxel aggregation function compared to a standard pooling operation. The attention layer can dynamically weight point features, which leads to a more expressive aggregation mechanism. In fact, the attention mechanism supersedes pooling operation, i.e., when the learned attention weight for each feature is one-hot based on the largest feature value, the attention module is equivalent to max pooling. Meanwhile, when the learned attention weight is the same for all the points, the attention module becomes the same as average pooling. 

\textbf{Latent query.}  A simple approach is using a randomly initialized latent vector, as Set Transformer and Perceiver \cite{lee2019set,jaegle2021perceiver}. Here, a single randomly initialized latent vector $Q\in \mathbb{R}^{1\times d}$ is used to query a set of LiDAR point features $L \in \mathbb{R}^{n \times d}$, where $d$ represents the number of features and $n$ represents the number of points inside a voxel. Point features are aggregated and encoded as a single voxel feature vector by $V = Transformer(q=Q, v=L, k=L) \in \mathbb{R}^{1 \times d}$, where the latent variable $Q$ are learned end-to-end during training. However, we find that using a latent query can cause training instability. 

\textbf{Residual query.} To address training instability, we propose using a residual query vector to stabilize the training. We use the max pooled features of the points within a voxel as a prior, which is added to the latent query. This residual query variable is used to query and encode the point features, leading to the final voxel representation, as shown in \autoref{fig:point_arch} (b). 

The residual variable $RQ = maxpool(L) + Q$, where $RQ \in \mathbb{R}^{1\times d}$, is used to query the point features, i.e.,  $V = Transformer(q=RQ, v=L, k=L) \in \mathbb{R}^{1 \times d}$. The latent query $Q$ allows a more flexible query vector representation than using max pooled feature alone, while the max pooled feature vector $maxpool(L)$ helps stabilize the training. 

\subsection{Model scaling}
\pva{} mitigates the point-voxel information bottleneck, which leads to the second contribution of this work: \textit{systematic study of the scaling properties of the Transformer-based 3D detectors}. To the best of our knowledge, this is the first time the scaling of Transformer-based 3D object detection has been explored in depth. Our results not only show that \pva{} has better-scaling properties by achieving state-of-the-art performance and providing valuable new insights on the best scaling practice. 

\textbf{Factorized scaling.} In this study, we analyze the scalability of Transformer-based 3D detectors by factorizing \pva{} architectures into two separate components: the point architecture and the voxel architecture, as depicted in \autoref{fig:arch}. Our \pva{} exhibits significantly better scaling properties compared to PointNet and proves to be a more effective solution than simply scaling the voxel architecture, which is prone to overfitting. 

\textbf{Data Augmentation.} Strong regularization is important when scaling up the model capacity. We use LidarAugment, including 10 data augmentations, during training \cite{leng2022lidaraugment}.

\textbf{Optimization for TPU accelerators.} Despite the recent advancements in optimizing 3D detectors for GPU training and inference, the use of TPU accelerators, which are widely adopted for training \textit{large-scale} Transformer models, has not been extensively explored for 3D detection models.

TPU accelerators are efficient for dense, static-shape computations, but less so for dynamic operations. In this study, we identify that dynamic voxelization proposed in \cite{zhou2020end} is 2X slower on TPU, despite being FLOPS efficient as shown in \autoref{tab:tpu_speed}, thus for the following experiments, we use fixed voxelization.

\section{Experiments}
Our primary evaluation of the \pva{} architecture is conducted on the Waymo Open Dataset \cite{sun2020scalability}, which comprises 798 training, 202 validation, and 150 test run segments, each run containing around 200 frames, including 360 degrees of LiDAR around the ego-vehicle.  The Waymo Open Dataset offers a unique advantage over other 3D outdoor detection datasets \cite{geiger2013vision,caesar2020nuscenes}, as it contains more than $10\times$ annotated Lidar frames, providing a valuable opportunity to study the scalability of transformer-based 3D detectors.

\subsection{Implementation details}

\begin{figure}[h!]
    \centering
    \vspace{0.1cm}
    \includegraphics[width=0.6\linewidth]{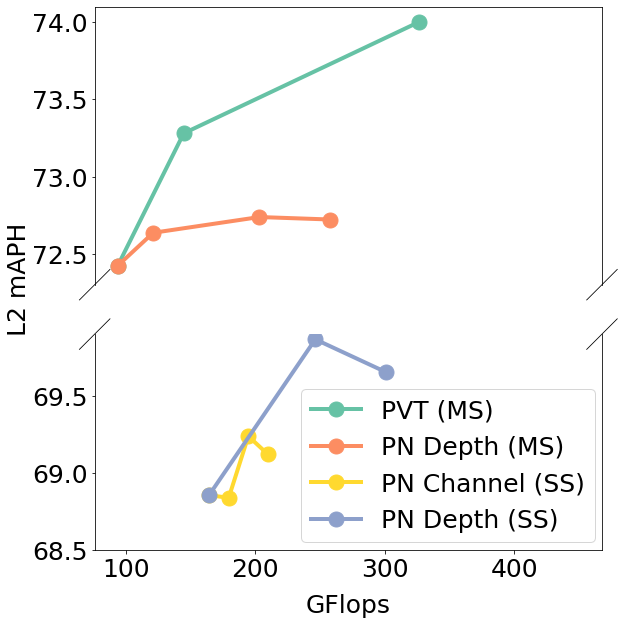}
    \caption{\textbf{\pva{}: better scalability.} Increasing PointNet's (PN) depth (red, purple) and channel (yellow) yields modest performance improvements, while scaling \pva{} PVT (green) shows significant improvements.  Previous works, both single-scale (SS) \cite{singlestride21} and multi-scale (MS) \cite{sun2022swformer} architectures, use PointNet for point feature aggregation, yet it underperforms when scaled beyond certain thresholds, leading to overfitting. \pva{} (green) overcomes these limitations by incorporating a Transformer-based point-to-voxel encoder, enabling effective scaling beyond 300 GFlops and achieving 74.0 mAPH L2 for vehicle and pedestrian detection on the Waymo Open Dataset Validation set.
    }
    \label{fig:point_scaling}
\end{figure}

\textbf{Overall architecture of \pva{}.}
The \pva{} architecture is composed of a point architecture and a voxel architecture, as shown in \autoref{fig:arch}. The key novelty of \pva{} lies in its point architecture, which utilizes a transformer module to aggregate point features. This is different from other PointNet-based transformer 3D detectors \cite{sun2022swformer, singlestride21}. The point architecture extracts point features and transforms them into $0.32\,$m $\times 0.32\,$m bird's eye view pillars, as shown in \autoref{fig:point_arch}. These features are then fed into the voxel architecture to predict bounding boxes. The box detection is through CenterNet heads \cite{yin2021centerpoint}. We follow the loss implementations in \cite{sun2022swformer}, which use a weighted average of foreground segmentation loss, box regression loss, and heatmap loss. We experimented with both single-scale and multiscale backbone designs and selected the multiscale design for its superior performance, as shown in \autoref{fig:point_scaling}.  

\textbf{Efficient static-shape point architecture.}  Dynamic voxelization \cite{zhou2020end} is widely used in 3D detection models due to its Flops efficiency. However, we have found that dense operations can be optimized through TPU XLA, leading to much more efficient computation, as demonstrated in  \autoref{tab:tpu_speed}. 

We benchmark the latency of a depth-two PointNet with a channel size of 128. Here, we use a 3-frame training setting, as in \autoref{tab:validation}. We denote the original number of point features, total number of points, and total number of voxels as $f = 4$, $m = 199600$, and $N=52000$ respectively. 

For the dynamic voxelization method, the initial point tensors  $\tilde{P}\in \mathbb{R}^{m\times f}$ are transformed into $P\in \mathbb{R}^{m\times 128}$  via two fully connected layers. The transformed point tensors are then aggregated into a voxel tensor $V\in \mathbb{R}^{N\times 128}$ trough dynamic gather in the forward pass and scatter in the backward pass. 

In contrast, the PointNet fixed voxelization method simplifies the implementation by bucketing the point tensors into $\tilde{P} \in \mathbb{R}^{N \times 32 \times f}$ and transforming them into  $P\in \mathbb{R}^{N \times 32 \times 128}$ through two fully connected layers. Here, we cap the maximum number of points per voxel to 32. The voxel features $V\in \mathbb{R}^{N\times 128}$ are then aggregated via a simple max pooling operation along the second axis of $P$, with all the operations being static without any dynamic gather or scatter. 

In our proposed \pva{} architecture, we take advantage of the simplification provided by the fixed voxelization approach.  First, the point features are transformed into $P\in \mathbb{R}^{N \times 32 \times 128}$ via a single fully connected layer. Then, points within each pillar bucket are treated as tokens, with a total sequence length of 32. We apply a point self-attention Transformer among point tokens within each voxel to further featurized points. Finally, point-to-voxel aggregation is implemented as a cross-attention Transformer between the point tensor $P\in \mathbb{R}^{N \times 32 \times 128}$  and a querying tensor $Q\in \mathbb{R}^{N \times 1 \times 128}$, along the second axis of $P$ and $Q$, resulting in a voxel tensor with shape $V\in \mathbb{R}^{N \times 1 \times 128}$. The implementation of the querying vector is shown in \autoref{fig:point_arch} (b). 

\begin{table}[]
 \centering
 \vspace{0.2cm}
 \resizebox{0.48\textwidth}{!}{
\begin{tabular}{c|cc|cc|cc}
\toprule
                     & \multicolumn{2}{c|}{batch size 1} & \multicolumn{2}{c|}{batch size 2} & \multicolumn{2}{c}{batch size 4} \\
\midrule
Latency (ms)         & forward        & backward        & forward        & backward        & forward        & backward        \\
\midrule
Dynamic voxelization & 13.7           & 16.6            & 24.3           & 33.1            & 48.3           & 66.0            \\
Fixed voxelization   & \textbf{6.5}            & \textbf{7.8}             &\textbf{ 12.1}           & \textbf{15.7  }          & \textbf{23.4}           & \textbf{31.3}\\
Speed up             & \textbf{2.1$\times$}           & \textbf{2.1$\times$ }           & \textbf{2.0$\times$}             & \textbf{2.1$\times$  }          & \textbf{2.1$\times$   }         &\textbf{ 2.1$\times$}            \\
\bottomrule
\end{tabular}
}
\caption{\textbf{2$\times$ training speed with fixed voxelization.}}
\label{tab:tpu_speed}
\end{table}

\textbf{Training setup.} We employ the AdamW \cite{loshchilov2017decoupled} optimizer and a cosine learning rate with warm up. The initial warm-up learning rate 3e-4, with a peak learning rate 1.2e-3, 10000 warm-up steps, and a total of 320000 training steps. The model is trained with a global batch size 64 for about 128 epochs. For most of the experiments, we use a 3-frame setting, stacking point clouds from the current frame, the previous frame, and the frame before the previous one.

\subsection{\pva{} for 3D object detection}
\autoref{tab:validation} and \autoref{tab:test_result} compares the performance of our \pva{} to other recent 3D detectors. The main comparisons for \pva{} are SWFormer and SST ~\cite{sun2022swformer, singlestride21}, both are transformer-based detectors and trained with the 3 frames as our \pva{}. Note, both SWFormer and SST are using a simple PointNet as the point architecture as opposed to our new attention design, as shown in \autoref{fig:arch}.

As shown in \autoref{tab:validation}, our \pva{} improves upon the state-of-the-art 3D detectors across all metrics, including L1 and L2 difficulties, the Vehicle and Pedestrian classes, and AP and APH. These performance gains are consistent in the test set as shown in \autoref{tab:test_result}, with an improvement of 1.7 and 1.9 mAPH L2 than recent state-of-the-art transformer detector SWFormer and convolution detector PillarNet \cite{shi2022pillarnet} and achieving a new state-of-the-art 76.5 mAPH L2 on the test set, demonstrating the effectiveness of our \pva{} architecture.

\begin{table}[!t]
\vspace{2mm}
\centering
 \resizebox{0.5\textwidth}{!}{
\begin{tabular}{l|c|cc|ccc}
\toprule
\multirow{2}{*}{Method} &
\multirow{1}{*}{mAPH} &
\multicolumn{2}{c|}{Vehicle AP/APH 3D} &
\multicolumn{2}{c}{Pedestrian AP/APH 3D} \\
& L2 & L1 & L2 & L1 & L2 \\
\midrule
P.Pillars~\cite{lang2019pointpillars}  & 51.9 & 63.3/62.7 &  55.2/54.7 &  68.9/56.6 & 60.4/49.1 \\
LiDAR R-CNN ~\cite{li2021LiDAR} &58.0 &73.5/73.0&64.7/64.2&71.2/58.8&63.1/51.7\\
CenterPoint~\cite{yin2021centerpoint}  & 67.1 & 76.6/76.1 & 68.9/68.4 & 79.0/73.4 & 71.0/65.8 \\
PVRCNN ~\cite{shi2020pvWOD} &67.3&77.5/76.9&68.7/68.2&78.9/75.1&69.8/66.4&\\
AFDetV2\cite{hu2022afdetv2}  & 68.1 & 77.6/77.1 & 69.7/69.2 & 80.2/74.6 & 72.2/67.0\\
RSN\_3f~\cite{sun2021rsn}   & 68.1 & 78.4/78.1 & 69.5/69.1 & 79.4/76.2 & 69.9/67.0 \\
PVRCNN++~\cite{shi2021pv++}   & 69.1 & 79.3/78.8 & 70.6/70.2 & 81.8/76.3 & 73.2/68.0  \\
SST\_3f \cite{singlestride21} & 69.5 &77.0/76.6 & 68.5/68.1 & 82.4/78.0 & 75.1/70.9 \\
UPillars-L\_3f \cite{leng2022lidaraugment} & {71.0}& {79.5/79.0} & {71.9/71.5} & {81.5/77.3} & {74.5/70.5} \\

PillarNet-34\_2f \cite{shi2022pillarnet}  & 71.8 &80.0/79.5 & 72.0/71.5 & 82.5/79.3 & 75.0/72.0 \\
% SWFormer (Original) \cite{sun2022swformer}  & 70.9 & 79.4/78.9 & 71.1/70.6 & 82.9/79.0 & 74.8/71.1 \\
CenterFormer\_2f \cite{zhou2022centerformer} & {72.5}& {77.0/76.5} & {72.1/71.6} & {81.4/78.0} & {76.7/73.4} \\
SWFormer\_3f$^\dagger$\cite{sun2022swformer}   & 72.8 & 80.9/80.4 & 72.8/72.4 & 84.4/80.7 & 76.8/73.2 \\
\midrule
\textbf{\pva{}\_3f}   &\textbf{74.0}& \textbf{81.9/81.3} & \textbf{73.8/73.4}& \textbf{85.3/81.8} &\textbf{78.0/74.6} \\
\textbf{\pva{}\_3f\_L}   &\textbf{74.4}& \textbf{82.2/81.7} & \textbf{74.3/73.9}& \textbf{85.3/81.9} &\textbf{78.2/74.8} \\
\bottomrule
\end{tabular}
}
\centering
\caption{\textbf{\pva{} outperforms other state-of-the-art architectures.} \pva{} increases all detection metrics compared to recent state-of-the-art SWFormer, CenterFormer, and PillarNet. PVT only scales the point architecture (see \autoref{tab:point_scaling_space}), while PVT\_L scales both point and voxel architectures (see \autoref{tab:bb_search_space}). Waymo Open Dataset validation set 3D detection metrics are reported. $^\dagger$ value reported in \cite{leng2022lidaraugment}. Our reproduced model has 72.6 mAPH L2. }
\label{tab:validation}
\end{table}

\begin{table}[!h]
\centering
\vspace{2mm}
 \resizebox{0.5\textwidth}{!}{
\begin{tabular}{l|c|cc|ccc}
\toprule
\multirow{2}{*}{Method} &
\multirow{1}{*}{mAPH} &
\multicolumn{2}{c}{Vehicle AP/APH 3D} &
\multicolumn{2}{c}{Pedestrian AP/APH 3D} \\
& L2 & L1 & L2 & L1 & L2 \\
\midrule
P.Pillars~\cite{lang2019pointpillars}  &55.1 & 68.6/68.1 &  60.5/60.1 &  68.0/55.5 & 61.4/50.1 \\
CenterPoint~\cite{yin2021centerpoint}  & 69.1 &80.2/79.7 & 72.2/71.8 & 78.3/72.1 & 72.2/66.4 \\
RSN\_3f~\cite{sun2021rsn} &69.7 & 80.7/80.3 & 71.9/71.6 & 78.9/75.6 & 70.7/67.8 \\
PVRCNN++~\cite{shi2021pv++} & 71.2  &81.6/81.2  & 73.9/73.5 & 80.4/75.0 & 74.1/69.0 \\
SST\_TS\_3f~\cite{singlestride21}  &72.9&81.0/80.6 & 73.1/72.7 & 83.1/79.4 & 76.7/73.1 \\
PillarNet-34\_2f \cite{shi2022pillarnet}  &74.6&83.2/82.8 & 76.1/75.7 & 82.4/79.0 & 76.7/73.5 \\
SWFormer\_3f \cite{sun2022swformer}  & 74.8 & 84.0/83.6 & 76.3/76.0 & 83.1/79.3 & 77.2/73.5 \\
\midrule
\textbf{\pva{}\_3f} &\textbf{76.1}& \textbf{84.8/84.5} & \textbf{77.3/77.0}& \textbf{84.2/80.6} & \textbf{78.5/75.1} \\
\textbf{\pva{}\_3f\_L} &\textbf{76.5}& \textbf{85.0/84.6} & \textbf{77.6/77.3}& \textbf{84.7/81.2} & \textbf{79.0/75.7} \\
\bottomrule
\end{tabular}
}
\centering
\caption{\textbf{New state-of-the-art test-set results on Waymo Open Dataset.} \pva{} achieves a new state-of-the-art mAPH L2,  with +1.7 mAPH gain compared to the prior art of SWFormer.}
\label{tab:test_result}
\vspace{-0.5cm}
\end{table}

\begin{figure}[h!]
    \centering
    \vspace{0.1cm}
    \includegraphics[width=1.0\linewidth]{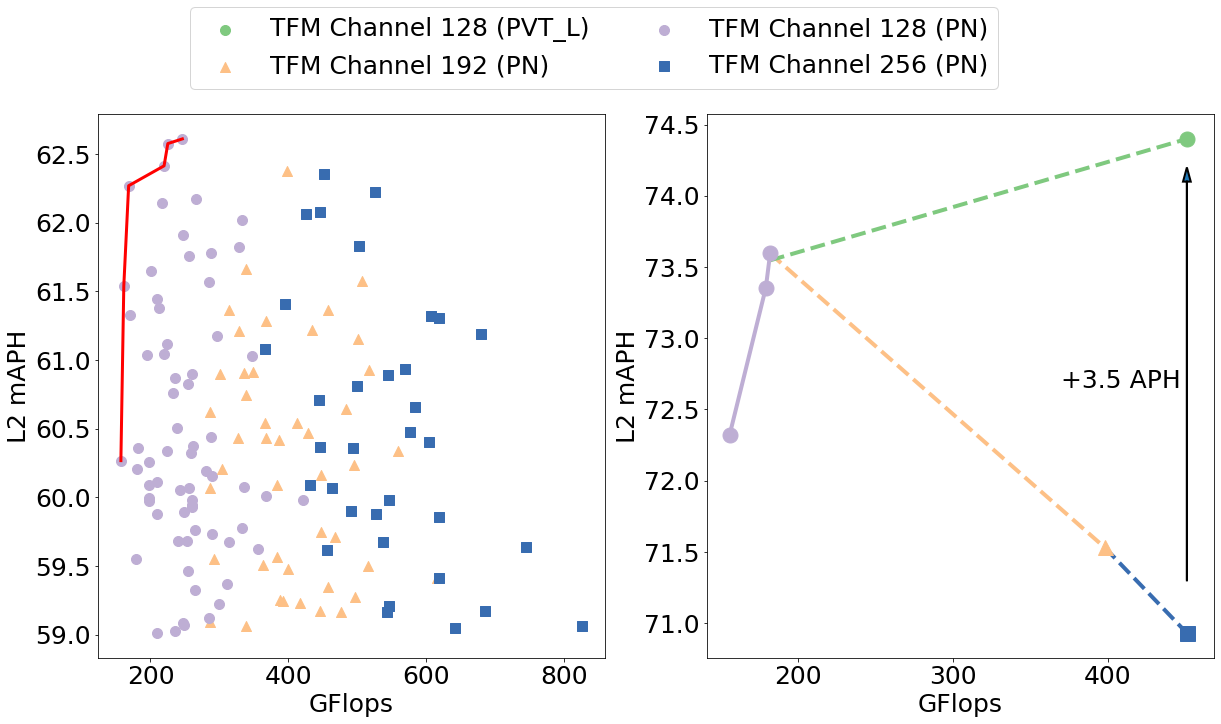}
    \caption{\textbf{The voxel architecture has limited scalability when using PointNet (PN) to aggregate point features.}  \textbf{Right:} Using Transformer to aggregate point features (PVT\_L) is significantly better (green) compared to using PointNet and scaling only the channels in the voxel architecture to 256 (blue), a 3.5 mAPH L2 increase at a similar Flops. \textbf{Left:} Performance of random sampled voxel architectures from the search space (in \autoref{tab:bb_search_space}) after trained for 12.8 epochs. We observe that scaling the voxel architecture while using PointNet can lead to suboptimal performance. The Pareto curve (red curve) shows scaling voxel architecture channels from 128 to 192 and 256 channels leads to overfittings. Waymo Open Dataset validation set mAPH L2 on Vehicle and Pedestrian are reported.}
    \label{fig:backbone_scaling}
\end{figure}
                 \begin{table}[]
\centering
\vspace{0.2cm}
\resizebox{0.35\textwidth}{!}{
\begin{tabular}{c|c}
\toprule
FC Channel                    & {[}\textcolor{orange}{\textbf{128}}, 192, 256, 320{]}  \\
PointNet FC Depth                      & {[}2, 5, 7{]}          \\
\pva{} Depth & {[}FC, FC-PV, \textcolor{orange}{\textbf{FC-PP-PV}}{]}\\
\bottomrule
\end{tabular}
}
\caption{\textbf{Point architecture and scaling space.} The point architecture for both \pva{}\_3f and \pva{}\_3f\_L are colored in orange. \textbf{PointNet.} Two fully connected (FC) layers with channel size 128 are used as the seed architecture. PointNet is scaled either by increasing the first layer FC channels or adding more FC layers. \textbf{\pva{}}, A single  FC layer is used as the seed point architecture. The \pva{} is scaled by adding a point-to-voxel aggregation layer (PV), as shown in \autoref{fig:point_arch}, and an additional point-to-point self-attention Transformer (PP) within the same voxel, as shown in \autoref{fig:arch}, \ref{fig:point_arch}. }
\label{tab:point_scaling_space}
\vspace{-4mm}
\end{table}
            
\subsection{Scaling Transformer-based 3D detector}
In this section, we present the first systematic architecture scaling study for Transformer-based 3D detector and demonstrate that our proposed \pva{} scales better compared to scaling \textit{point architectures} (PointNet) and \textit{voxel architectures} (window Transformer) proposed in prior works. These results emphasize the significance of our novel Transformer-based point encoding module. 

\subsubsection{Scaling point architecture}
The point architecture in \pva{} encodes sparse point features into grid-like voxel features via Transformer architecture, shown in \autoref{fig:arch}. The key difference between \pva{} and prior works lies in learning to aggregate point features through an attention mechanism rather than a heuristic pooling operation. This approach in \pva{} alleviates the point-to-voxel information bottleneck introduced by PointNet, which has to aggregate information from over 30+ points through a simple max pooling.

\textbf{Point architecture scaling space.}
We explore the scalability of the point architecture by scaling up our proposed \pva{} architecture and the conventional PointNet architecture. The scaling space for point architecture is summarized in \autoref{tab:point_scaling_space}.  The seed point architecture for PointNet is composed of two fully connected layers with hidden channels of [128, 128], as described in \cite{sun2022swformer}. We explore two scaling strategies for PointNet, namely increasing the hidden channel size of the first layer or stacking more fully connected layers, as shown in \autoref{tab:point_scaling_space}.

For \pva{}, the seed architecture consists of a fully connected layer to featurize the point feature to 128 dimensions, followed by the proposed point-aggregation module, as shown in \autoref{fig:point_arch}. To scale up the point architecture of \pva{}, we introduce an additional point-point self-attention layer between the fully connected layer and the point-aggregation module, as shown in \autoref{fig:point_arch}.

In addition, we investigate the point architecture scaling using two types of Transformer backbones, a single-scale backbone and a multi-scale backbone. For the single-scale backbone, we follow \cite{singlestride21} and use 6 layers of window Transformer at the same resolution 0.32$\,$m. For multiscale backbone, we follow \cite{sun2022swformer} and use 12 layers of window Transformer blocks with [2, 3, 2, 3, 2] layers at a resolution [0.32$\,$m, 0.64$\,$m, 1.28$\,$m, 5.12$\,$m, 10.24$\,$m]. For both backbones, we use the same hidden channel of 128, 8 heads, and an expansion ratio of 2 for all Transformer layers.

\textbf{Superior scalability of \pva{}.}
As shown in \autoref{fig:point_scaling}, our \pva{} demonstrates better scaling compared to PointNet. Our multi-scale \pva{} improves upon single-layer PointNet from 72.4 mAPH L2 to 74.0 mAPH L2, a 1.6 mAPH increase. Notably, the scaling performance is not achievable with commonly used PointNet, as increasing the depth or hidden channel size leads to performance plateau and overfitting, as evidenced by the saturation and decline in performance beyond a certain scale.

\subsubsection{Scaling voxel architecture}
The voxel architecture predicts bounding boxes based on voxelized features, which include both the backbone and detection head architectures. To capture the intertwined nature of different components in the voxel architecture, we parameterize the voxel architectures, as shown in \autoref{tab:bb_search_space}, and random sample 150 architectures from the search space to evaluate the performance. To save cost, we train each architecture on the full training set but reduce the training steps to 1/10, a total of 12.8 epochs as a proxy setting.  Here, we use two-layer PointNet as the point architecture. Our results show without the proposed \pva{}, scaling voxel architecture is suboptimal. 
\begin{table}[]
 \vspace{0.2cm}
\resizebox{0.49\textwidth}{!}{
\begin{tabular}{c|c||c|c}
\toprule
\multicolumn{2}{c||}{Backbone}        & \multicolumn{2}{c}{Head}                \\
\midrule
TFM Channel       & {[}\textcolor{Orange}{\textbf{128}}, 192, 256{]} & TFM Channel           & {[}\textcolor{Orange}{\textbf{128}}, 192, 256{]} \\
TFM Heads         & {[}4, \textcolor{NavyBlue}{\textbf{8}}, \textcolor{red}{\textbf{16}}{]}      & TFM Heads             & {[}\textcolor{red}{\textbf{4}}, \textcolor{NavyBlue}{\textbf{8}}, 16{]}      \\
TFM MLP Expansion     & {[}\textcolor{Orange}{\textbf{2}}, 4{]}          & TFM MLP Expansion         & {[}\textcolor{NavyBlue}{\textbf{2}}, \textcolor{red}{\textbf{4}}{]}          \\
0.32$\,$m Num of Blocks  & {[}\textcolor{NavyBlue}{\textbf{2}}, 3, \textcolor{red}{\textbf{4}}, 6{]}       & Vehicle Num of Blocks    & {[}\textcolor{Orange}{\textbf{1}}, 2, 4, 6{]}    \\
0.64$\,$m Num of Blocks  & {[}2, \textcolor{NavyBlue}{\textbf{3}}, \textcolor{red}{\textbf{4}}, 6{]}       & Pedestrian Num of Blocks & {[}\textcolor{NavyBlue}{\textbf{1}}, \textcolor{red}{\textbf{2}}, 4, 6{]}    \\
1.28$\,$m Num of Blocks  & {[}\textcolor{NavyBlue}{\textbf{2}}, 3, \textcolor{red}{\textbf{4}}, 6{]}       &                   &                     \\
5.12$\,$m Num of Blocks  & {[}2, \textcolor{NavyBlue}{\textbf{3}}, \textcolor{red}{\textbf{4}}, 6{]}       &                   &                     \\
10.24$\,$m Num of Blocks & {[}\textcolor{Orange}{\textbf{\textbf{2}}}, 3, 4, 6{]}       &                   &                 \\
\bottomrule
\end{tabular}
}
\caption{\textbf{Voxel architecture and search space.} The parameters for \pva{}\_3f and \pva{}\_3f\_L backbones are highlighted in blue and red respectively. The value is colored orange when both backbones share the same parameter. We search the hyperparameters for each Transformer layer, including the channel size, number of heads, and fully connected layer expansion ratio for both the backbone and detection heads. In addition, we search the number of repeated blocks for each Transformer layer at each resolution and for each detection head. }
\label{tab:bb_search_space}
\vspace{-0.2cm}
\end{table}

\textbf{Voxel architecture scaling is challenging.} Our architecture studies, as shown in \autoref{fig:backbone_scaling}, reveal that the model performance plateaus after scaling up the voxel architecture beyond 250 GFlops. Interestingly, we find that scaling Transformer channel size does not lead to better model performance. The performance-computation Pareto frontier strictly lies in the channel size 128 architecture regime. 

Upon retraining the selected voxel architectures on the Pareto frontier as well as two additional voxel architectures with Transformer channel sizes 192 and 256 for 128 epochs, we find wider voxel architectures are prone to overfitting, while a smaller 128-channel model has the best performance with 73.5 mAPH L2. 

We further show our \pva{} is more scalable by replacing PointNet with our attention aggregation module and scaling the architecture to 452 GFlops. Our PVT\_L achieves a much higher 74.4 mAPH L2, a 3.5 mAPH increase, compared to scaling the Transformer channel size to 256 at a similar 451 GFLops, as shown in \autoref{fig:teaser}, \ref{fig:backbone_scaling}.

\subsection{Generalize to the single frame setting}
In addition to 3-frame temporal training, \autoref{tab:single_frame} shows our \pva{} significantly improves the single-frame setting, leading to a +3.0 mAPH improvement.

\begin{table}[!t]
\vspace{2mm}
\centering
 \resizebox{0.48\textwidth}{!}{
\begin{tabular}{l|c|cc|ccc}
\toprule
\multirow{2}{*}{Method} &
\multirow{1}{*}{mAPH} &
\multicolumn{2}{c|}{Vehicle AP/APH 3D} &
\multicolumn{2}{c}{Pedestrian AP/APH 3D} \\
& L2 & L1 & L2 & L1 & L2 \\
\midrule
PointNet\_1f  & 68.1 & 79.0/78.5 & 70.5/70.1 & 81.5/73.5 & 73.5/66.1 \\
\textbf{\pva{}\_1f}   &\textbf{71.1}& \textbf{80.9/80.4} & \textbf{72.5/72.1}& \textbf{83.7/77.6} &\textbf{75.9/70.1} \\
\bottomrule
\end{tabular}
}
\centering
\caption{\textbf{\pva{} generalizes to single frame settings.} WOD val. set 3D detection metrics are reported.}
\label{tab:single_frame}
\end{table}
\subsection{Ablation on different aggregation methods}
As shown in \autoref{tab:aggregation}, our \pva{} also outperforms averaged pooling based feature aggregation, demonstrating that naive averaging is suboptimal and learning to aggregate, as proposed in this work, is necessary.
\begin{table}[!t]
\vspace{2mm}
\centering
 \resizebox{0.48\textwidth}{!}{
\begin{tabular}{l|c|cc|ccc}
\toprule
\multirow{2}{*}{Method} &
\multirow{1}{*}{mAPH} &
\multicolumn{2}{c|}{Vehicle AP/APH 3D} &
\multicolumn{2}{c}{Pedestrian AP/APH 3D} \\
& L2 & L1 & L2 & L1 & L2 \\
\midrule
PN averaged pool\_3f  &  72.1& 80.1/79.7 & 72.2/71.8& 83.7/79.8 & 76.1/72.4 \\
PN max pool\_3f  & 72.8 & 80.9/80.4 & 72.8/72.4& 84.4/80.7 & 76.8/73.2\\
\textbf{\pva{}\_3f}   &\textbf{74.0}& \textbf{81.9/81.3} & \textbf{73.8/73.4}& \textbf{85.3/81.8} &\textbf{78.0/74.6} \\
\bottomrule
\end{tabular}
}
\centering
\caption{\textbf{\pva{} outperforms other point aggregation methods.} WOD val. set 3D detection metrics are reported.}
\label{tab:aggregation}
\end{table}

\section{Conclusion}
In this work, we aim to enable  better scalability for large-scale 3D object detectors and identify that the pooling-based PointNet introduces an information bottleneck for modern 3D object detectors. To address this limitation, we propose a new \pva{} architecture, which uses an attention-based Transformer to aggregate point features into voxel features. We show that such a point-to-voxel Transformer is more expressive than a simple PointNet's pooling layer, and thus achieve much better performance than previous 3D object detectors. Our \pva{} significantly outperforms the prior art, such as SWFormer, and achieves new state-of-the-art results on the challenging Waymo Open Dataset.

{\small
\bibliographystyle{ieee_fullname}
\bibliography{egbib}

\begin{thebibliography}{10}\itemsep=-1pt

\bibitem{bewley2020range}
Alex Bewley, Pei Sun, Thomas Mensink, Dragomir Anguelov, and Cristian Sminchisescu.
\newblock Range conditioned dilated convolutions for scale invariant 3d object detection.
\newblock In {\em Conference on Robot Learning}, 2020.

\bibitem{caesar2020nuscenes}
Holger Caesar, Varun Bankiti, Alex~H Lang, Sourabh Vora, Venice~Erin Liong, Qiang Xu, Anush Krishnan, Yu Pan, Giancarlo Baldan, and Oscar Beijbom.
\newblock nuscenes: A multimodal dataset for autonomous driving.
\newblock In {\em Proceedings of the IEEE/CVF conference on computer vision and pattern recognition}, pages 11621--11631, 2020.

\bibitem{chai2021point}
Yuning Chai, Pei Sun, Jiquan Ngiam, Weiyue Wang, Benjamin Caine, Vijay Vasudevan, Xiao Zhang, and Dragomir Anguelov.
\newblock To the point: Efficient 3d object detection in the range image with graph convolution kernels.
\newblock In {\em Proceedings of the IEEE/CVF Conference on Computer Vision and Pattern Recognition}, pages 16000--16009, 2021.

\bibitem{singlestride21}
Lue Fan, Ziqi Pang, Tianyuan Zhang, Yu-Xiong Wang, Hang Zhao, Feng Wang, Naiyan Wang, and Zhaoxiang Zhang.
\newblock Embracing single stride 3d object detector with sparse transformer.
\newblock In {\em Proceedings of the IEEE/CVF conference on computer vision and pattern recognition}, pages 8458--8468, 2022.

\bibitem{fan2021rangedet}
Lue Fan, Xuan Xiong, Feng Wang, Naiyan Wang, and Zhaoxiang Zhang.
\newblock Rangedet: In defense of range view for lidar-based 3d object detection.
\newblock In {\em Proceedings of the IEEE/CVF International Conference on Computer Vision}, pages 2918--2927, 2021.

\bibitem{geiger2013vision}
Andreas Geiger, Philip Lenz, Christoph Stiller, and Raquel Urtasun.
\newblock Vision meets robotics: The kitti dataset.
\newblock {\em The International Journal of Robotics Research}, 32(11):1231--1237, 2013.

\bibitem{hu2022afdetv2}
Yihan Hu, Zhuangzhuang Ding, Runzhou Ge, Wenxin Shao, Li Huang, Kun Li, and Qiang Liu.
\newblock Afdetv2: Rethinking the necessity of the second stage for object detection from point clouds.
\newblock In {\em Proceedings of the AAAI Conference on Artificial Intelligence}, volume~36, pages 969--979, 2022.

\bibitem{jaegle2021perceiver}
Andrew Jaegle, Felix Gimeno, Andy Brock, Oriol Vinyals, Andrew Zisserman, and Joao Carreira.
\newblock Perceiver: General perception with iterative attention.
\newblock In {\em International conference on machine learning}, pages 4651--4664. PMLR, 2021.

\bibitem{lang2019pointpillars}
Alex~H Lang, Sourabh Vora, Holger Caesar, Lubing Zhou, Jiong Yang, and Oscar Beijbom.
\newblock Pointpillars: Fast encoders for object detection from point clouds.
\newblock In {\em Proceedings of the IEEE/CVF conference on computer vision and pattern recognition}, pages 12697--12705, 2019.

\bibitem{lee2019set}
Juho Lee, Yoonho Lee, Jungtaek Kim, Adam Kosiorek, Seungjin Choi, and Yee~Whye Teh.
\newblock Set transformer: A framework for attention-based permutation-invariant neural networks.
\newblock In {\em International conference on machine learning}, pages 3744--3753. PMLR, 2019.

\bibitem{leng2022lidaraugment}
Zhaoqi Leng, Guowang Li, Chenxi Liu, Ekin~Dogus Cubuk, Pei Sun, Tong He, Dragomir Anguelov, and Mingxing Tan.
\newblock Lidaraugment: Searching for scalable 3d lidar data augmentations.
\newblock {\em arXiv preprint arXiv:2210.13488}, 2022.

\bibitem{li2021LiDAR}
Zhichao Li, Feng Wang, and Naiyan Wang.
\newblock Lidar r-cnn: An efficient and universal 3d object detector.
\newblock In {\em Proceedings of the IEEE/CVF Conference on Computer Vision and Pattern Recognition}, pages 7546--7555, 2021.

\bibitem{liu2019point2sequence}
Xinhai Liu, Zhizhong Han, Yu-Shen Liu, and Matthias Zwicker.
\newblock Point2sequence: Learning the shape representation of 3d point clouds with an attention-based sequence to sequence network.
\newblock In {\em Proceedings of the AAAI Conference on Artificial Intelligence}, volume~33, pages 8778--8785, 2019.

\bibitem{liu2019point}
Zhijian Liu, Haotian Tang, Yujun Lin, and Song Han.
\newblock Point-voxel cnn for efficient 3d deep learning.
\newblock {\em Advances in Neural Information Processing Systems}, 32, 2019.

\bibitem{loshchilov2017decoupled}
Ilya Loshchilov and Frank Hutter.
\newblock Decoupled weight decay regularization.
\newblock {\em arXiv preprint arXiv:1711.05101}, 2017.

\bibitem{Meyer_2019_lasernet}
Gregory~P. Meyer, Ankit Laddha, Eric Kee, Carlos Vallespi-Gonzalez, and Carl~K. Wellington.
\newblock Lasernet: An efficient probabilistic 3d object detector for autonomous driving.
\newblock In {\em Proceedings of the IEEE/CVF Conference on Computer Vision and Pattern Recognition (CVPR)}, June 2019.

\bibitem{misra2021end}
Ishan Misra, Rohit Girdhar, and Armand Joulin.
\newblock An end-to-end transformer model for 3d object detection.
\newblock In {\em Proceedings of the IEEE/CVF International Conference on Computer Vision}, pages 2906--2917, 2021.

\bibitem{ngiam2019starnet}
Jiquan Ngiam, Benjamin Caine, Wei Han, Brandon Yang, Yuning Chai, Pei Sun, Yin Zhou, Xi Yi, Ouais Alsharif, Patrick Nguyen, et~al.
\newblock Starnet: Targeted computation for object detection in point clouds.
\newblock {\em arXiv preprint arXiv:1908.11069}, 2019.

\bibitem{qi2018frustum}
Charles~R Qi, Wei Liu, Chenxia Wu, Hao Su, and Leonidas~J Guibas.
\newblock Frustum pointnets for 3d object detection from rgb-d data.
\newblock In {\em Proceedings of the IEEE conference on computer vision and pattern recognition}, pages 918--927, 2018.

\bibitem{qi2017pointnet}
Charles~R Qi, Hao Su, Kaichun Mo, and Leonidas~J Guibas.
\newblock Pointnet: Deep learning on point sets for 3d classification and segmentation.
\newblock In {\em Proceedings of the IEEE conference on computer vision and pattern recognition}, pages 652--660, 2017.

\bibitem{qi2017pointnet++}
Charles~Ruizhongtai Qi, Li Yi, Hao Su, and Leonidas~J Guibas.
\newblock Pointnet++: Deep hierarchical feature learning on point sets in a metric space.
\newblock {\em Advances in neural information processing systems}, 30, 2017.

\bibitem{shi2022pillarnet}
Guangsheng Shi, Ruifeng Li, and Chao Ma.
\newblock Pillarnet: High-performance pillar-based 3d object detection.
\newblock {\em arXiv preprint arXiv:2205.07403}, 2022.

\bibitem{shi2020pvWOD}
Shaoshuai Shi, Chaoxu Guo, Jihan Yang, and Hongsheng Li.
\newblock Pv-rcnn: The top-performing lidar-only solutions for 3d detection/3d tracking/domain adaptation of waymo open dataset challenges.
\newblock {\em arXiv preprint arXiv:2008.12599}, 2020.

\bibitem{shi2021pv++}
Shaoshuai Shi, Li Jiang, Jiajun Deng, Zhe Wang, Chaoxu Guo, Jianping Shi, Xiaogang Wang, and Hongsheng Li.
\newblock Pv-rcnn++: Point-voxel feature set abstraction with local vector representation for 3d object detection.
\newblock {\em arXiv preprint arXiv:2102.00463}, 2021.

\bibitem{shi2019pointrcnn}
Shaoshuai Shi, Xiaogang Wang, and Hongsheng Li.
\newblock Pointrcnn: 3d object proposal generation and detection from point cloud.
\newblock In {\em Proceedings of the IEEE/CVF conference on computer vision and pattern recognition}, pages 770--779, 2019.

\bibitem{sun2020scalability}
Pei Sun, Henrik Kretzschmar, Xerxes Dotiwalla, Aurelien Chouard, Vijaysai Patnaik, Paul Tsui, James Guo, Yin Zhou, Yuning Chai, Benjamin Caine, et~al.
\newblock Scalability in perception for autonomous driving: Waymo open dataset.
\newblock In {\em Proceedings of the IEEE/CVF conference on computer vision and pattern recognition}, pages 2446--2454, 2020.

\bibitem{sun2022swformer}
Pei Sun, Mingxing Tan, Weiyue Wang, Chenxi Liu, Fei Xia, Zhaoqi Leng, and Dragomir Anguelov.
\newblock Swformer: Sparse window transformer for 3d object detection in point clouds.
\newblock {\em arXiv preprint arXiv:2210.07372}, 2022.

\bibitem{sun2021rsn}
Pei Sun, Weiyue Wang, Yuning Chai, Gamaleldin Elsayed, Alex Bewley, Xiao Zhang, Cristian Sminchisescu, and Dragomir Anguelov.
\newblock Rsn: Range sparse net for efficient, accurate lidar 3d object detection.
\newblock In {\em Proceedings of the IEEE/CVF Conference on Computer Vision and Pattern Recognition}, pages 5725--5734, 2021.

\bibitem{vaswani2017attention}
Ashish Vaswani, Noam Shazeer, Niki Parmar, Jakob Uszkoreit, Llion Jones, Aidan~N Gomez, {\L}ukasz Kaiser, and Illia Polosukhin.
\newblock Attention is all you need.
\newblock In {\em NeurIPS}, 2017.

\bibitem{yang2019modeling}
Jiancheng Yang, Qiang Zhang, Bingbing Ni, Linguo Li, Jinxian Liu, Mengdie Zhou, and Qi Tian.
\newblock Modeling point clouds with self-attention and gumbel subset sampling.
\newblock In {\em Proceedings of the IEEE/CVF conference on computer vision and pattern recognition}, pages 3323--3332, 2019.

\bibitem{yin2021centerpoint}
Tianwei Yin, Xingyi Zhou, and Philipp Krahenbuhl.
\newblock Center-based 3d object detection and tracking.
\newblock In {\em Proceedings of the IEEE/CVF conference on computer vision and pattern recognition}, pages 11784--11793, 2021.

\bibitem{zhao2021point}
Hengshuang Zhao, Li Jiang, Jiaya Jia, Philip~HS Torr, and Vladlen Koltun.
\newblock Point transformer.
\newblock In {\em Proceedings of the IEEE/CVF International Conference on Computer Vision}, pages 16259--16268, 2021.

\bibitem{zhou2019end}
Yin Zhou, Pei Sun, Yu Zhang, Dragomir Anguelov, Jiyang Gao, Tom Ouyang, James Guo, Jiquan Ngiam, and Vijay Vasudevan.
\newblock End-to-end multi-view fusion for 3d object detection in lidar point clouds.
\newblock In {\em CORL}, 2019.

\bibitem{zhou2020end}
Yin Zhou, Pei Sun, Yu Zhang, Dragomir Anguelov, Jiyang Gao, Tom Ouyang, James Guo, Jiquan Ngiam, and Vijay Vasudevan.
\newblock End-to-end multi-view fusion for 3d object detection in lidar point clouds.
\newblock In {\em Conference on Robot Learning}, pages 923--932. PMLR, 2020.

\bibitem{zhou2018voxelnet}
Yin Zhou and Oncel Tuzel.
\newblock Voxelnet: End-to-end learning for point cloud based 3d object detection.
\newblock In {\em Proceedings of the IEEE conference on computer vision and pattern recognition}, pages 4490--4499, 2018.

\bibitem{zhou2022centerformer}
Zixiang Zhou, Xiangchen Zhao, Yu Wang, Panqu Wang, and Hassan Foroosh.
\newblock Centerformer: Center-based transformer for 3d object detection.
\newblock In {\em Computer Vision--ECCV 2022: 17th European Conference, Tel Aviv, Israel, October 23--27, 2022, Proceedings, Part XXXVIII}, pages 496--513. Springer, 2022.

\end{thebibliography}
}

\end{document}